\pdfoutput=1

\documentclass[11pt]{article}

\usepackage{ACL2023}

\usepackage{times}
\usepackage{latexsym}
\usepackage{graphicx} 
\usepackage{subcaption}
\usepackage{booktabs} 
\usepackage{multirow} 
\usepackage{enumitem} 
\usepackage{hyperref}

\usepackage[T1]{fontenc}

\usepackage[utf8]{inputenc}
\usepackage{microtype}

\usepackage{inconsolata}
\usepackage{amsmath}
\usepackage{cleveref}

\graphicspath{{img/}}

\title{When Cultures Move: Measuring and Improving \\ Multicultural Text-to-Video Generation}

\author{Shuowei Li \quad Yuming Zhao \quad Parth Bhalerao  \quad Oana Ignat \\
  Department of Computer Science and Engineering \\
  Santa Clara University, Santa Clara, CA, USA \\
  \texttt{\{yzhao4,pbhalerao,oignat\}@scu.edu} \\}

\begin{document}
\maketitle

\begin{abstract}
Text-to-video (T2V) generation has achieved strong visual fidelity, yet its
ability to represent multiple cultures within a single scene remains largely
unexplored, a gap with real consequences as generative video becomes a
mainstream medium for cultural storytelling. We introduce multicultural
text-to-video generation as a new task and present the first benchmark
designed to study this setting: 243 culturally grounded prompts and 972
videos, spanning three cultures (Chinese, American, Romanian), three action
categories, and both mono-cultural and cross-cultural scenarios. Using this
benchmark, we analyze how current T2V models represent culture with
CLIP-based metrics, VLM-as-judge assessments, and video quality measures. As
one strategy for improving cultural fidelity, we explore \textsc{MAVEN},
which decomposes prompts into \textit{person}, \textit{action}, and
\textit{location} dimensions, each handled by a culturally specialized agent
in parallel or sequentially. In our controlled setting, parallel specialization achieves the strongest cultural relevance while maintaining visual quality and temporal consistency, suggesting that structured prompt refinement is a promising direction for multicultural T2V generation. We release our dataset and
code at: \url{https://github.com/AIM-SCU/MAVEN}.
\end{abstract}

\section{Introduction}
Text-to-video (T2V) generation has rapidly advanced 
\cite{DBLP:conf/iclr/YangTZ00XYHZFYZ25,DBLP:journals/corr/abs-2405-10674,sora_system_card}, 
shifting the central challenge from visual realism alone to \emph{semantic faithfulness}. 
Among these dimensions, \emph{cultural grounding}, how people, actions, and places are 
represented with respect to specific cultures, remains both critically important and 
insufficiently understood. While prior work has documented systematic cultural gaps in 
language and image models 
\cite{DBLP:journals/corr/abs-2404-10199,DBLP:journals/corr/abs-2501-01282,DBLP:conf/nips/KannenAAPPDBD24}, 
the video setting, where cultural content must remain coherent across both space and time, 
has received almost no dedicated study. To our knowledge, this work is among the first
to systematically frame \emph{multicultural T2V generation} as a problem in its own right.

\begin{figure}[h]
\centering
\includegraphics[width=1.0\linewidth]{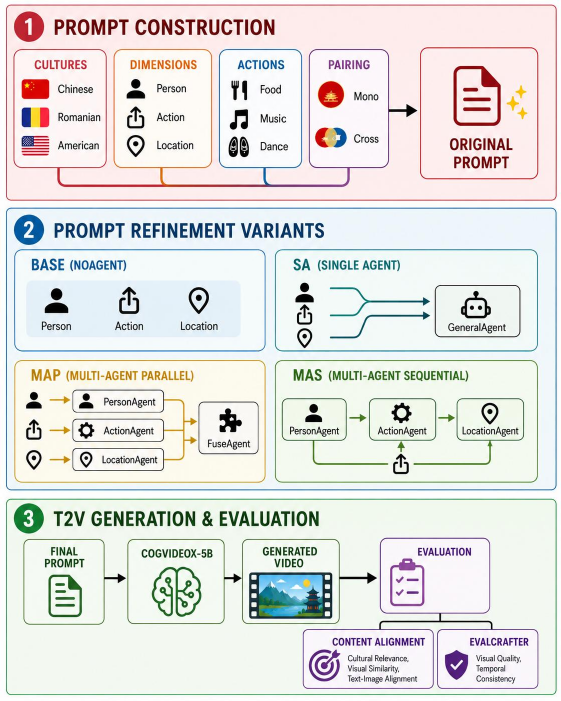}
\caption{Most T2V systems assume a single culture per scene, but real scenes are
often multicultural, combining person, action, and location from different
backgrounds. We explore agent-based prompt refinement, shown here, as one
strategy for this.}
\label{fig:main_figure}
\end{figure}

Culture is inherently compositional: a person from one cultural background may perform 
an action associated with another culture in a location tied to a third. Such 
\emph{cross-cultural} scenarios parallel the composition problem studied in image 
transcreation \cite{DBLP:conf/naacl/KhanujaIHN25} and matter for downstream uses such 
as inclusive content creation and accessibility. Yet current T2V benchmarks overwhelmingly 
assume mono-cultural prompts 
\cite{DBLP:journals/corr/abs-2411-13503,DBLP:journals/corr/abs-2507-18107}, implicitly 
treating culture as a single uniform attribute, and most T2V systems convey all cultural 
information through a single prompt or refinement agent, even though person appearance, 
action execution, and location depiction each draw on distinct forms of cultural expertise. 
Building on recent multi-agent T2V frameworks 
\cite{DBLP:journals/corr/abs-2403-13248,DBLP:journals/corr/abs-2508-08487}, we introduce 
\textsc{MAVEN} (\underline{M}ulti-\underline{A}gent \underline{V}ideo \underline{E}nrichment 
for cultural \underline{N}arrative), which decomposes cultural grounding into three core 
dimensions, \emph{person}, \emph{action}, and \emph{location}, each handled by a 
culturally specialized agent that enriches prompts before video generation. We evaluate 
on a benchmark of 243 prompts and 972 videos spanning three cultures, three action 
categories, and both mono-cultural and cross-cultural configurations 
(Section~\ref{sec:dataset}), using CLIP-based metrics 
\cite{DBLP:conf/icml/RadfordKHRGASAM21}, VLM judgments of cultural relevance 
\cite{DBLP:journals/corr/abs-2501-01282}, and video quality assessments 
\cite{DBLP:conf/cvpr/LiuC0WZCLZCS24}.

Beyond model performance, this problem carries real \textbf{societal stakes}: as T2V becomes a mainstream medium for storytelling, education, and marketing, faithfully representing cultures matters for avoiding stereotyping, erasure of underrepresented traditions, and misrepresentation, motivating our focus on \emph{measuring} cultural fidelity alongside improving it. Concretely, we organize our investigation around three research questions:
\begin{itemize}[noitemsep,topsep=2pt,leftmargin=1.5em]
    \item \textbf{RQ1.} Can multi-agent prompt refinement substantively improve cultural 
    fidelity in T2V generation, and does it close the gap between mono-cultural and 
    cross-cultural prompts?
    \item \textbf{RQ2.} How does the choice of agent communication structure, specifically 
    parallel specialization versus sequential refinement, affect cultural fidelity, visual 
    quality, and temporal consistency?
    \item \textbf{RQ3.} To what extent do CLIP-based automatic metrics agree with 
    VLM-based judgments of cultural relevance, and what does any disagreement reveal 
    about the limits of current T2V evaluation?
\end{itemize}

\section{Related Work}

\paragraph{Cultural Understanding in LLMs and VLMs.}
Cultural gaps have been documented in both LLMs \cite{DBLP:journals/corr/abs-2404-10199,DBLP:conf/nips/LiC0S024} 
and vision--language models \cite{DBLP:conf/nips/KannenAAPPDBD24, DBLP:journals/corr/abs-2501-01282,Nayak2025CulturalFrames}, 
pointing to systematic knowledge gaps driven by sparse geographic coverage in training data. 
Culturally grounded evaluation has been extended to multilingual visual question answering 
\cite{Romero2024CVQA} and to multicultural caption and image generation, where 
multi-agent frameworks with cultural personas have been shown to improve cross-cultural 
depiction in static images \cite{bai-etal-2025-power, bhalerao-etal-2026-cultures}. Our work shares these concerns but identifies two further limitations. First, cultures are
studied independently, with no existing work modeling cross-cultural \emph{composition}, i.e.,
combining person, action, and location from distinct cultural sources within a single
generation. Second, all prior work focuses on static image understanding or generation,
leaving the temporal consistency challenges unique to video entirely unaddressed.

\paragraph{Multi-Agent Text-to-Video Generation.}
Multi-agent frameworks for T2V generation decompose generation into specialized subtasks, 
including end-to-end agent coordination \cite{DBLP:journals/corr/abs-2403-13248}, film-crew 
simulation \cite{DBLP:journals/corr/abs-2408-11788,DBLP:journals/corr/abs-2503-07314}, 
cross-shot protagonist consistency \cite{DBLP:journals/corr/abs-2411-04925}, and 
script-writing alignment \cite{DBLP:journals/corr/abs-2508-08487}. A related line of work 
pursues LLM-guided prompt refinement for T2V along non-cultural axes 
\cite{Xue2024PhyT2V,Ji2024PromptAVideo,Gao2025RAPO,Yang2026SCMAPR}. We follow these 
frameworks in decomposing generation into specialized subtasks, but reorient agent roles 
from \emph{production} (director, editor, keyframe artist) to \emph{cultural dimensions} 
(person, action, location), enabling systematic study of both mono- and cross-cultural 
fidelity within a single prompt.

\paragraph{Cultural Text-to-Video Benchmarks.}
Existing benchmarks for cultural T2V evaluation cover scenario-level cultural fairness 
\cite{DBLP:journals/corr/abs-2411-13503,DBLP:conf/iclr/WangH00YML0C00024}, geo-cultural 
bias in city landscapes \cite{CaliskanIskakovLiMcCollumRen2025SimCityNet}, 
attribute-object-action compositionality \cite{DBLP:conf/cvpr/SunHL0XLL25}, and deep 
cultural knowledge integration \cite{DBLP:journals/corr/abs-2507-18107}. However, none address \emph{cross-cultural} composition, where person, action, and location originate from distinct cultures in a single system. Our benchmark directly targets this gap.

\section{Dataset}
\label{sec:dataset}

A central contribution of this paper is a new benchmark for culturally grounded T2V evaluation.
The benchmark comprises 243 unique prompts (Table~\ref{tab:dataset_stats}). Passing each through the four refinement pipelines (Base, SA, MAS, MAP; Section~\ref{sec:methodology}) yields 972 generated videos.
The cross-cultural split is intentionally twice the mono-cultural split, since cross-cultural composition is the central novelty of our benchmark and the regime where current T2V models fail most clearly.

\paragraph{Cultures and actions.}
We span three cultures (\textbf{Chinese}, \textbf{American}, \textbf{Romanian}) chosen for geographic and linguistic distinctness, and three action categories (\textbf{food}, \textbf{music}, \textbf{dance}) chosen for visual distinctiveness and the existence of clearly filmable cultural variants.
Each (culture $\times$ category) pair contributes 3 actions and each culture contributes 3 landmarks. The full list is given in Table~\ref{tab:culture_pal}.

\paragraph{Prompt construction.}
Prompts follow the template ``$P$ $A$ at $L$'' (person, action, location).
\textbf{Mono-cultural} prompts fix all three roles to a single culture, yielding $3 \times 3 \times 3 \times 3 = 81$ prompts.
\textbf{Cross-cultural} prompts draw $P$, $A$, $L$ from three \emph{distinct} cultures
($c_p \neq c_a \neq c_l$), yielding $3! \times (3 \times 3) \times 3 = 162$ prompts.
A representative example is \textit{``an American person eating dumplings at Bran Castle''} ($P$=American, $A$=Chinese, $L$=Romanian), which requires rendering three distinct cultural sources within a single video.

\begin{table}[t]
\centering
\small
\begin{tabular}{lr}
\toprule
\textbf{Component} & \textbf{Count} \\
\midrule
Cultures (Chinese, American, Romanian)            & 3   \\
Action categories (food, music, dance)            & 3   \\
Actions per (culture $\times$ category)           & 3   \\
Distinct actions                                  & 27  \\
Landmarks per culture                             & 3   \\
Distinct landmarks                                & 9   \\
\midrule
Mono-cultural prompts                             & 81  \\
\textbf{Cross-cultural prompts (novel)}           & \textbf{162} \\
\textbf{Total unique prompts}                     & \textbf{243} \\
\midrule
Refinement pipelines (Base, SA, MAS, MAP)         & 4   \\
\textbf{Total generated videos}                   & \textbf{972} \\
Duration per video (seconds)                      & 5   \\
\textbf{Total video duration (minutes)}           & \textbf{81}  \\
\bottomrule
\end{tabular}
\caption{MAVEN benchmark statistics.}
\label{tab:dataset_stats}
\end{table}

\begin{table}[t]
\centering
\small
\setlength{\tabcolsep}{4pt}
\renewcommand{\arraystretch}{0.95}
\begin{tabular}{lp{5.2cm}}
\toprule
\textbf{Category} & \textbf{Items} \\
\midrule
\multicolumn{2}{l}{\textbf{Chinese}} \\
Food      & Peking duck, mooncakes, dumplings \\
Music     & guzheng, erhu, dizi \\
Dance     & fan dance, ribbon dance, umbrella dance \\
Locations & Forbidden City, West Lake, Potala Palace \\
\addlinespace
\multicolumn{2}{l}{\textbf{American}} \\
Food      & hot dogs, burgers, pizza slice \\
Music     & banjo, electric guitar, saxophone \\
Dance     & hip-hop, moonwalk, tap dance \\
Locations & Statue of Liberty, Grand Canyon, Mount Rushmore \\
\addlinespace
\multicolumn{2}{l}{\textbf{Romanian}} \\
Food      & sarmale, mici, m\u{a}m\u{a}lig\u{a} \\
Music     & nai, cob\u{z}\u{a}, \c{t}ambal \\
Dance     & Hora, S\^{a}rba, Br\^{a}ul \\
Locations & Bran Castle, Palace of Parliament, Wooden Churches of Maramure\c{s} \\
\bottomrule
\end{tabular}
\caption{Full list of cultural items in the MAVEN benchmark. Actions in prompts are formed by prepending the canonical verb (\emph{eating}, \emph{playing}, \emph{dancing}) to each Food/Music/Dance item.}
\label{tab:culture_pal}
\end{table}

\section{Methodology}
\label{sec:methodology}

Our methodology covers three components: (1) agent-based prompt refinement pipelines, (2) a fixed T2V generation model to ensure fair comparison across pipelines, and (3) implementation details for reproducibility.
Figure~\ref{fig:main_figure} provides an overview.

\subsection{Agent-Based Prompt Refinement}

While base prompts include explicit cultural markers, they often lack the fine-grained
visual and contextual details needed for culturally faithful video generation.
We therefore introduce agent-based prompt refinement, guided by the hypothesis that
cultural knowledge for person appearance, action execution, and location depiction
belongs to distinct domains.

Each agent is assigned a culture-specific persona corresponding to the dimension it
refines: for example, an ActionAgent refining a Chinese food action is instructed as a
culturally grounded observer of Chinese dining practices.
Each agent's system prompt combines a dynamically generated cultural persona with a
dimension-specific instruction; concrete prompts are provided in
Appendix~\ref{sec:appendix_prompts}.

\paragraph{Pipeline designs.}
We compare four prompt refinement pipelines, all taking an original prompt $\text{Pr}_{\text{orig}}$ and producing a final prompt $\text{Pr}_{\text{final}}$.

\textbf{Base (No-Agent).}
\[
\text{Pr}_{\text{final}} = \text{Pr}_{\text{orig}}
\]

\textbf{Single-Agent (SA).}
A single general-purpose agent jointly refines all prompt dimensions using a unified
instruction and persona:
\[
\text{Pr}_{\text{final}} = \text{SingleAgent}(\text{Pr}_{\text{orig}}).
\]

\textbf{Multi-Agent Parallel (MAP).}
Three specialist agents independently refine the prompt in parallel, each targeting a
single dimension:
\begin{align*}
\text{Pr}_{P} &= \text{PersonAgent}(\text{Pr}_{\text{orig}}), \\
\text{Pr}_{A} &= \text{ActionAgent}(\text{Pr}_{\text{orig}}), \\
\text{Pr}_{L} &= \text{LocationAgent}(\text{Pr}_{\text{orig}}).
\end{align*}
A fusion agent then merges their outputs into a coherent prompt:
\[
\text{Pr}_{\text{final}} = \text{FuseAgent}([\text{Pr}_{P},\, \text{Pr}_{A},\, \text{Pr}_{L}]).
\]

\textbf{Multi-Agent Sequential (MAS).}
The same three specialist agents refine the prompt sequentially, each operating on the
previous agent's output:
\begin{align*}
\text{Pr}_{P} &= \text{PersonAgent}(\text{Pr}_{\text{orig}}), \\
\text{Pr}_{A} &= \text{ActionAgent}(\text{Pr}_{P}), \\
\text{Pr}_{\text{final}} &= \text{LocationAgent}(\text{Pr}_{A}).
\end{align*}

These pipelines allow us to compare general versus specialized refinement, as well as
parallel versus sequential agent coordination.

\subsection{Text-to-Video Model}
\label{subsec:t2v_model}

All videos are generated using CogVideoX-5B \cite{DBLP:conf/iclr/YangTZ00XYHZFYZ25}, an open-source diffusion-based text-to-video model.
We use a fixed generation setup across all experiments (5-second videos at $720\times480$ resolution, 8 fps, 50 inference steps, guidance scale 6, fixed seed), ensuring that observed differences are attributable solely to prompt refinement strategies rather than model variation.
\href{https://github.com/THUDM/CogVideo}{CogVideoX-5B} offers a strong balance between generation quality, computational cost, and reproducibility, and directly supports text-only prompts, making it well suited to our framework.

\subsection{Implementation Details}
\label{subsec:implementation}

Prompt refinement is implemented using a unified agent interface, with all agent calls logged for reproducibility.

All pipelines are executed asynchronously, enabling parallel agent execution where applicable.
All agents are instantiated using LLaMA-3.1-70B \cite{DBLP:journals/corr/abs-2407-21783}, served locally via Ollama and orchestrated through the AutoGen multi-agent framework \cite{DBLP:journals/corr/abs-2308-08155}.

\section{Evaluation and Results}
\label{sec:eval_results}

\subsection{Evaluation Metrics}
\label{subsec:eval_metrics}

We evaluate generated videos along three complementary dimensions: cultural relevance, visual similarity, and text--image alignment. These dimensions are chosen to capture both whether cultural content is faithfully represented and whether prompt refinement introduces meaningful visual change or simply rewrites surface form.

\paragraph{Cultural Relevance.}
We define a Cultural Relevance Score (CRS) using CLIP \cite{DBLP:conf/icml/RadfordKHRGASAM21} embeddings.
For each prompt, we construct four cultural grounding statements (CGS) targeting distinct dimensions: \textbf{OCGS} (``This image belongs to \{country\}.''), \textbf{PCGS} (``This image shows \{person description\}.''), \textbf{ACGS} (``This image depicts \{action\}, a practice associated with \{culture\} culture.''), and \textbf{LCGS} (``This image shows \{landmark\} in \{country\}.'').
For each video, we uniformly sample 5 frames across the video duration and compute frame-level cosine similarity with each CGS using CLIP:
\[
\text{CRS}(v, \text{CGS}) = \frac{1}{5}\sum_{i=1}^{5} \text{sim}(f_i, \text{CGS}).
\]
The four dimension scores --- OCRS (overall), PCRS (person), ACRS (action), and LCRS (location) --- are averaged to yield the overall CRS.

\paragraph{Visual Similarity.}
We compute a Visual Similarity Score (VSS) to measure how much refinement alters visual content.
For each prompt, we pair the base video $v_{\text{base}}$ with an agent-refined video $v_{\text{agent}}$, uniformly sample 5 corresponding frames across each video, and compute frame-wise CLIP similarity:
\[
\text{VSS}(v) = \frac{1}{5}\sum_{i=1}^{5} \text{sim}(f_{\text{base},i},\, f_{\text{agent},i}).
\]
A high VSS indicates that refinement enriches cultural and semantic details without substantially altering the overall scene layout, which we verify in the results below.

\paragraph{Text--Image Alignment.}
We compute a frame-level CLIP alignment score $\text{AS}(v, t)$ between the frames of a generated video $v$ and a text description $t$, where $t$ is the original prompt $\text{Pr}_{\text{orig}}$, the refined prompt $\text{Pr}_{\text{final}}$, or a cultural grounding statement.
Two derived metrics quantify the effect of refinement: \textbf{cultural enrichment}, the gain in alignment with the \emph{original} prompt for the agent video versus the base video; and \textbf{cultural relevance improvement}, the gain in alignment with the cultural grounding statements. However, because agent refinement also produces longer and more detailed prompts, our current design does not fully isolate the effect of cultural specialization from prompt elaboration; a detail-matched non-cultural refinement baseline would help disentangle these effects.

\subsection{VLM-Based Evaluation}
\label{subsec:vlm_eval}
To complement automatic metrics, we use a vision--language model (Gemini 2.5 Pro) \cite{DBLP:journals/corr/abs-2507-06261} as a judge \cite{DBLP:journals/corr/abs-2501-01282}.
The VLM evaluates cultural relevance, visual similarity, and text--image alignment using the middle frame of the five uniformly sampled frames from each video, providing a consistent frame-level evaluation across all videos. Scores range from 1 (not culturally relevant) to 5 (highly culturally relevant) and are aggregated across dimensions to obtain VLM-based counterparts to CRS and VSS. Since the VLM observes a single representative frame, these scores assess frame-level cultural representation rather than temporal consistency across the full video. Exact evaluation prompts are provided in Appendix~\ref{subsec:appendix_vlm_prompts}.

\subsection{Cultural Relevance Results}
\begin{figure*}[t]
\centering
\includegraphics[width=0.62\textwidth]{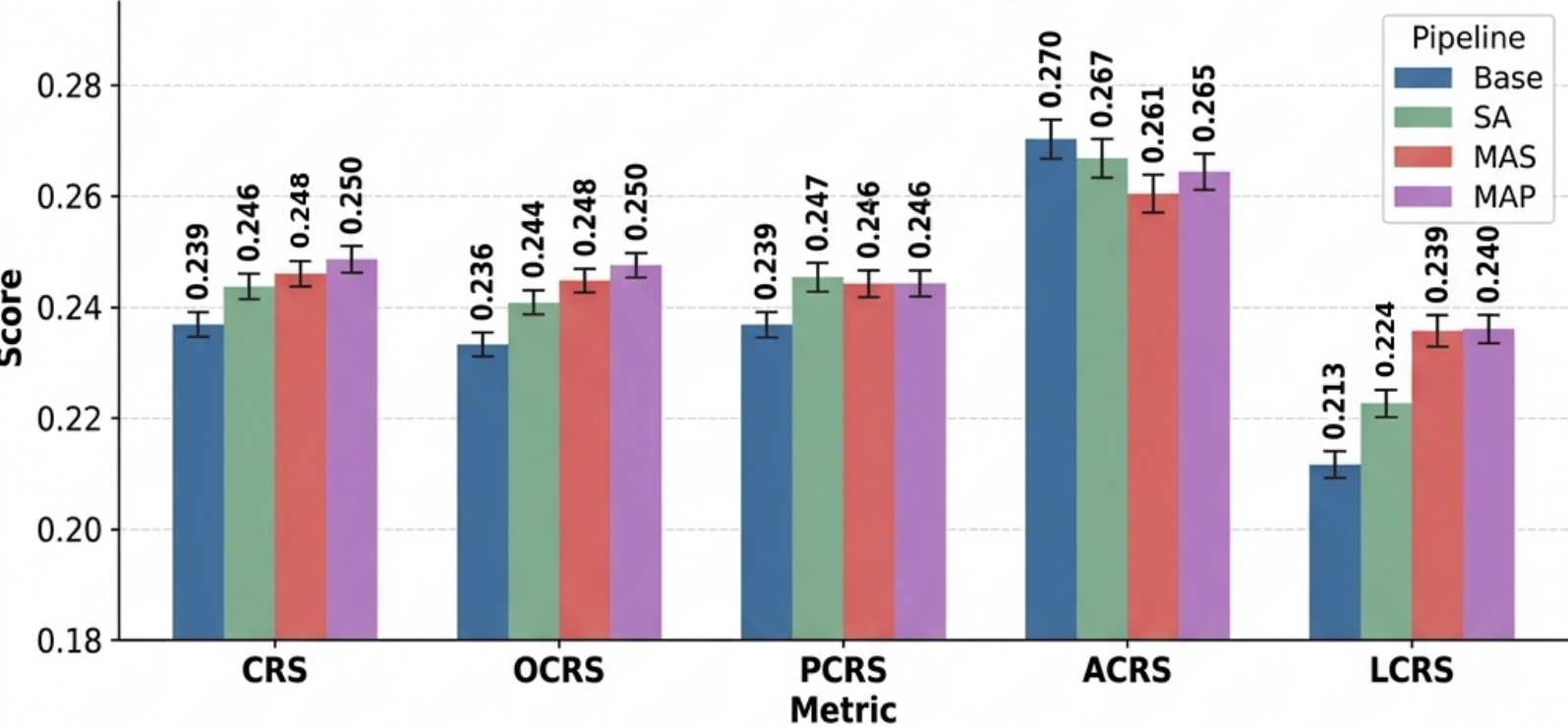}
\caption{CRS and dimension-specific scores (OCRS, PCRS, ACRS, LCRS) for all four pipelines. Error bars show 95\% CIs. MAP achieves the highest overall CRS (0.250), with statistically significant gains over Base (no-agent baseline) on CRS and LCRS.}
\label{fig:crs_results}
\end{figure*}
Multi-agent pipelines outperform both the base and single-agent (SA) baselines on the Cultural Relevance Score (CRS), with the parallel variant (MAP) achieving the highest overall CRS (Figure~\ref{fig:crs_results}).

\paragraph{Overall performance.}
MAP improves CRS by +4.6\% over the base pipeline and +1.6\% over SA, with MAS following closely.
These results suggest that distributing refinement across specialized agents
can yield stronger cultural grounding than a single general agent in our setting.

\paragraph{Dimension-level analysis.}
Improvements are not uniform across dimensions. Person-related relevance improves
consistently, though appearance traits occupy a narrow visual footprint and are less
reliably captured by frame-level embedding similarity. Action relevance stays stable,
as action cues are inherently more ambiguous for CLIP and richer prompts introduce
competing visual elements. Location shows the largest gains, with multi-agent
pipelines improving LCRS by over 12\% relative to Base, more than double SA's gain,
since architectural features translate more directly into CLIP-measurable signals.
Overall, MAP achieves the most balanced improvements, supporting the potential
benefit of dimension-specific specialization for cultural fidelity in our setting.

\begin{figure}[h]
\centering
\includegraphics[width=\linewidth]{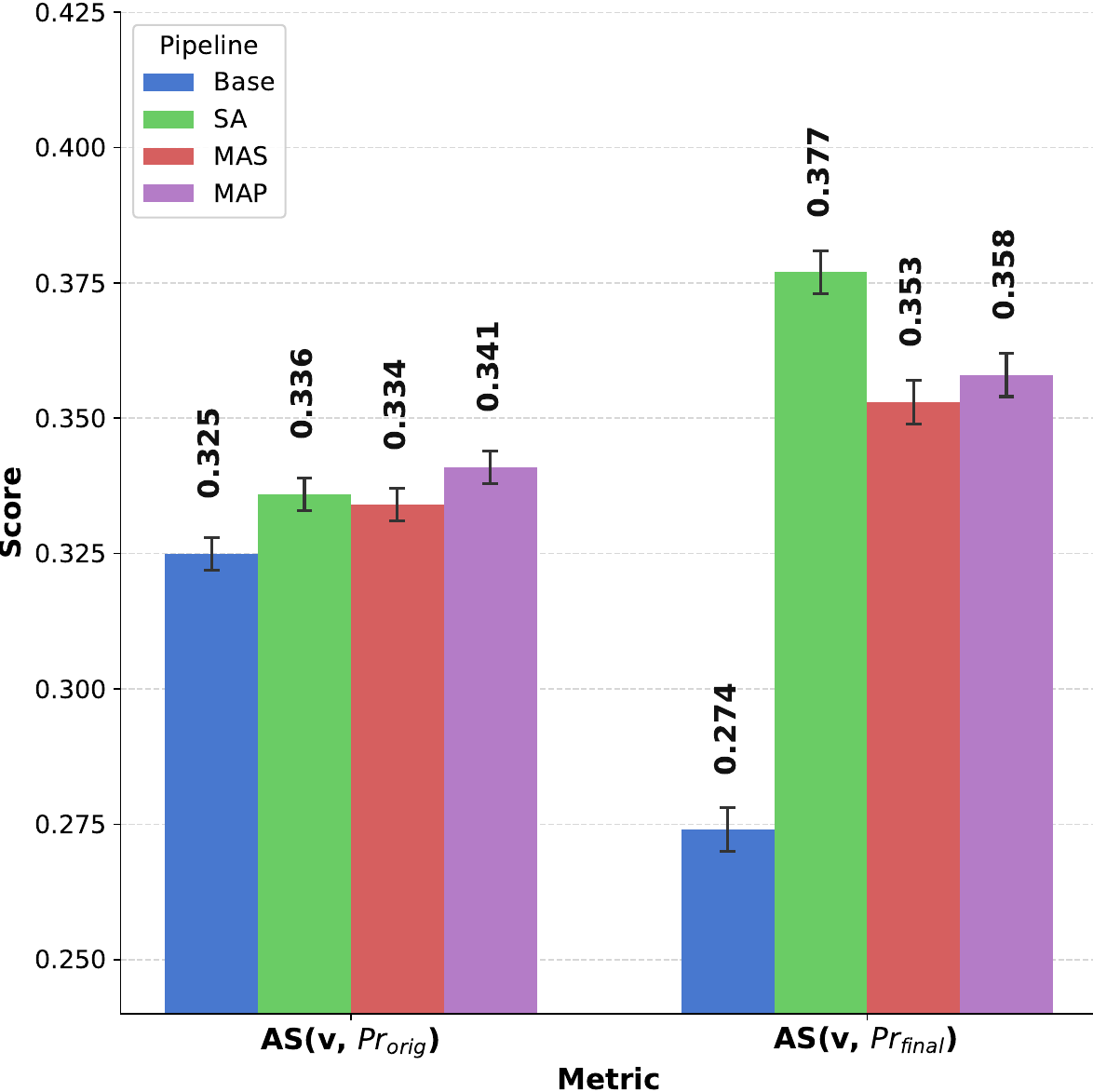}
\caption{Alignment scores for all four pipelines with the original prompt $\text{Pr}_{\text{orig}}$ and refined prompt $\text{Pr}_{\text{final}}$. Error bars show 95\% CIs. All proposed methods (SA, MAS, MAP) improve significantly over Base (no-agent baseline) on refined prompt alignment, with MAP achieving the highest original prompt alignment (0.341).}
\label{fig:enrichment_results}
\vspace{-1.0em}
\end{figure}



\subsection{Visual Similarity}
Despite improving cultural relevance, agent-based pipelines preserve high visual similarity
to base videos (VSS $>$ 0.68 for all pipelines; differences among SA, MAS, and MAP are
$<$1.2\%). This indicates that refinement primarily enriches semantic and cultural details
rather than altering overall scene layout, which is the intended behavior: culturally
faithful generation should deepen the cultural grounding of a scene, not reconstruct it
from scratch.

\subsection{Text--Image Alignment Results}
\label{subsec:text_image_alignment}
Agent-based refinement improves alignment with both the original and refined prompts
(Figure~\ref{fig:enrichment_results}).
\paragraph{Alignment with original prompts.}
MAP achieves the strongest cultural enrichment, improving alignment by +4.9\% over base.
This indicates that refinement helps T2V models better express the cultural intent already
present in the original prompt, rather than drifting from it.
\paragraph{Alignment with refined prompts.}
All agent pipelines dramatically improve alignment with refined prompts (+29\% to +38\%), reflecting the increased specificity of enriched prompts.
SA achieves the highest alignment here, likely due to greater stylistic coherence from a single-agent rewrite.
However, alignment with the original prompt, which is more indicative of cultural fidelity, remains highest for MAP, reinforcing its practical advantage.

\subsection{Mono vs.\ Cross-Cultural Results}
Cross-cultural prompts are consistently more challenging than mono-cultural ones, with 
the average CRS decreasing by approximately 4 to 6\% across all pipelines. The one 
exception is OCRS, which scores higher for cross-cultural prompts because it rewards 
explicit cultural fusion, whereas the per-dimension scores (PCRS, ACRS, LCRS) penalize 
the visual incoherence that arises when multiple cultures coexist within a single 
generated frame. Agent-based refinement narrows this gap, with MAP achieving the 
smallest per-dimension performance drop, suggesting that parallel specialization is 
particularly effective in cross-cultural scenarios.

\begin{figure*}[t]
\centering
\includegraphics[width=0.62\textwidth]{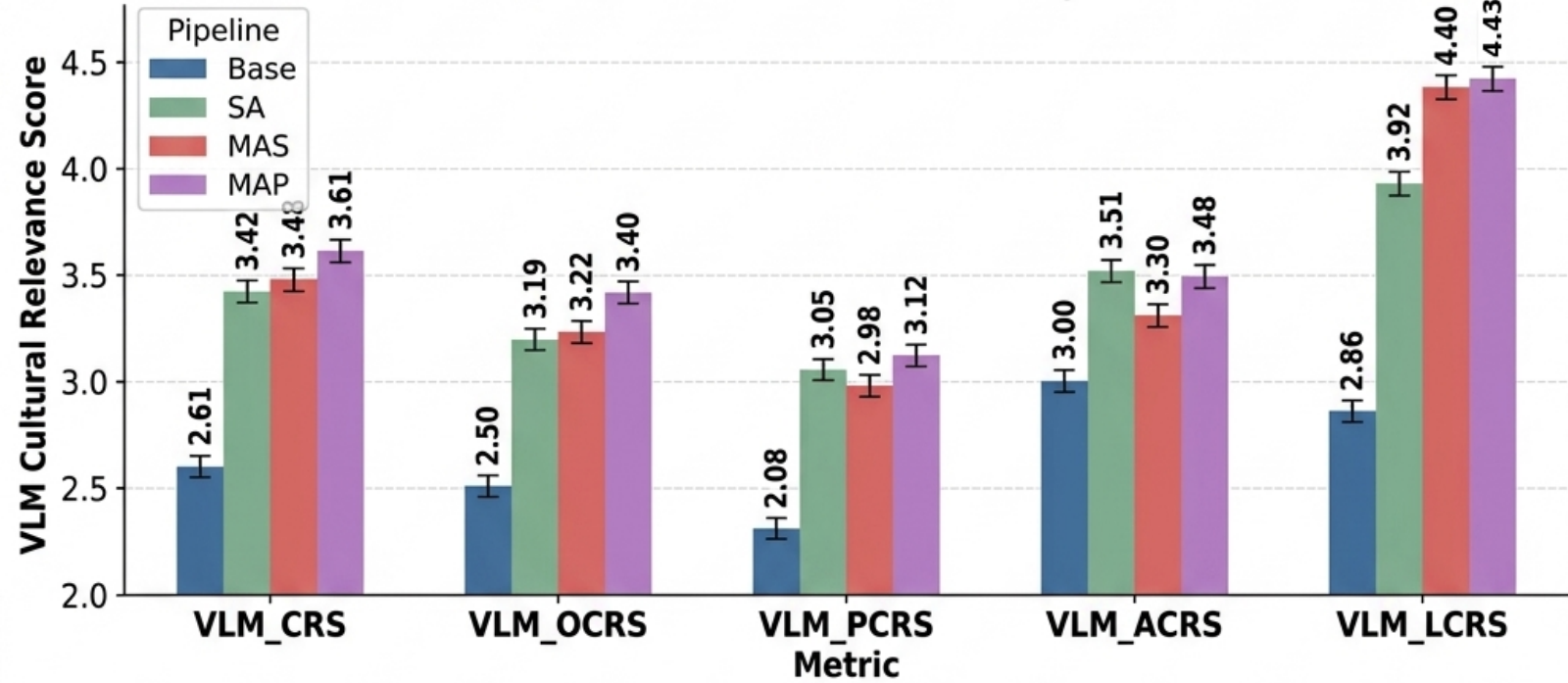}
\caption{VLM-judged cultural relevance scores (scored 1--5) for all four pipelines 
across overall and dimension-specific metrics. Error bars show 95\% CIs. MAP achieves 
the highest VLM\_CRS (3.61), with statistically significant improvements over Base 
(no-agent baseline) across all dimensions, most notably on VLM\_LCRS 
(Base: 2.86 $\to$ MAP: 4.43).}
\label{fig:vlm_crs_results}
\end{figure*}

\subsection{VLM-Judged Cultural Relevance Results}
VLM-based evaluation (Gemini 2.5 Pro) independently corroborates CLIP-based findings
(Figure~\ref{fig:vlm_crs_results}).
MAP achieves the highest VLM-judged cultural relevance (VLM\_CRS = 3.61),
improving over base by +38\% and over SA by +5.6\%.
Location remains the strongest dimension (VLM\_LCRS: base 2.86 $\to$ MAP 4.43, +54.9\%),
while person-related cues remain the most challenging (VLM\_PCRS: base 2.08 $\to$ MAP 3.12),
indicating persistent difficulty in modeling culturally specific appearance.

\subsection{VLM-Judged Visual Similarity Results}
VLM-judged visual similarity scores (2.05--2.09 on a 1--5 scale) reveal a
divergence from CLIP-based VSS: while CLIP reports high similarity ($>$0.68),
VLM assigns substantially lower scores.
This gap indicates that agent refinement introduces culturally meaningful visual changes
that are perceptually salient to a semantic judge but not captured by embedding distance.
All three pipelines score similarly, suggesting that the degree of visual change
is not strongly dependent on the specific agent architecture in our setting.

\subsection{VLM-Judged Text--Image Alignment Results}
VLM-judged alignment further confirms MAP's advantage in enriching original prompt meaning,
with a +61.3\% improvement over base (MAP: 3.00 vs.\ base: 1.86).
Alignment with refined prompts is highest for SA (2.88), again reflecting stylistic
coherence from single-agent rewriting rather than cultural depth.
The large gap between alignment with the original prompt (MAP: 3.00) and the refined
prompt (base: 1.04) confirms that enriched prompts contain substantially richer cultural
content than base videos represent.

\subsection{Metric Correlation Results}
CLIP-based \cite{DBLP:conf/icml/RadfordKHRGASAM21} and VLM-judged metrics show moderate to strong Pearson correlation on person, action, and location, validating CLIP as a useful proxy: PCRS is most consistent (r = 0.61--0.66), followed by ACRS (r = 0.40--0.53) and LCRS (r = 0.40--0.71). Correlation is near zero for the overall dimension (OCRS), likely due to its abstract country-level framing. The base pipeline reaches the highest correlation on most dimensions, while agent-refined pipelines drop substantially: on LCRS, base reaches r = 0.71 versus 0.40--0.52 for agent pipelines, a 27--43\% relative decrease. This suggests refinement introduces nuanced cultural cues better captured by VLM reasoning than embedding similarity.

\subsection{Video Quality and Temporal Consistency Results}
\begin{figure}[h]
\centering
\includegraphics[width=0.85\linewidth]{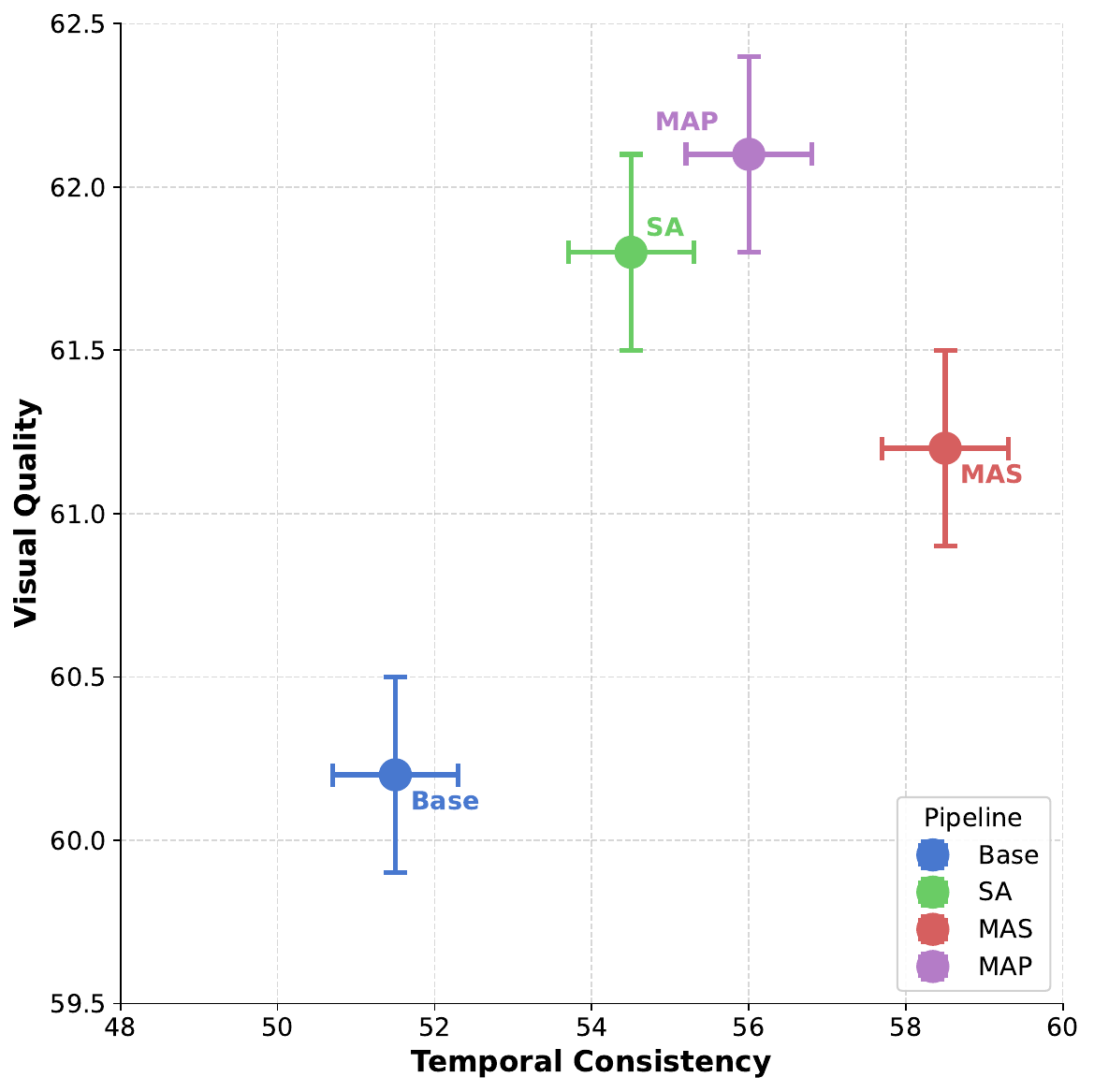}
\caption{Visual Quality vs.\ Temporal Consistency for all four pipelines. Each point 
represents one pipeline; error bars show 95\% CIs. All proposed methods improve over 
Base (no-agent baseline) on both metrics. SA leads on visual quality, MAS on temporal 
consistency, and MAP achieves the most balanced improvement across both.}
\label{fig:video_quality_scatter}
\vspace{-0.9em}
\end{figure}
Agent-based refinement improves both visual quality and temporal consistency
(Figure~\ref{fig:video_quality_scatter}).
While SA maximizes visual quality and MAS maximizes temporal consistency, MAP achieves the
most balanced improvement across both metrics.
This balance, combined with its superior cultural relevance, makes MAP the most robust
overall refinement strategy.

\begin{figure*}[!t]
\centering
\includegraphics[width=0.74\textwidth]{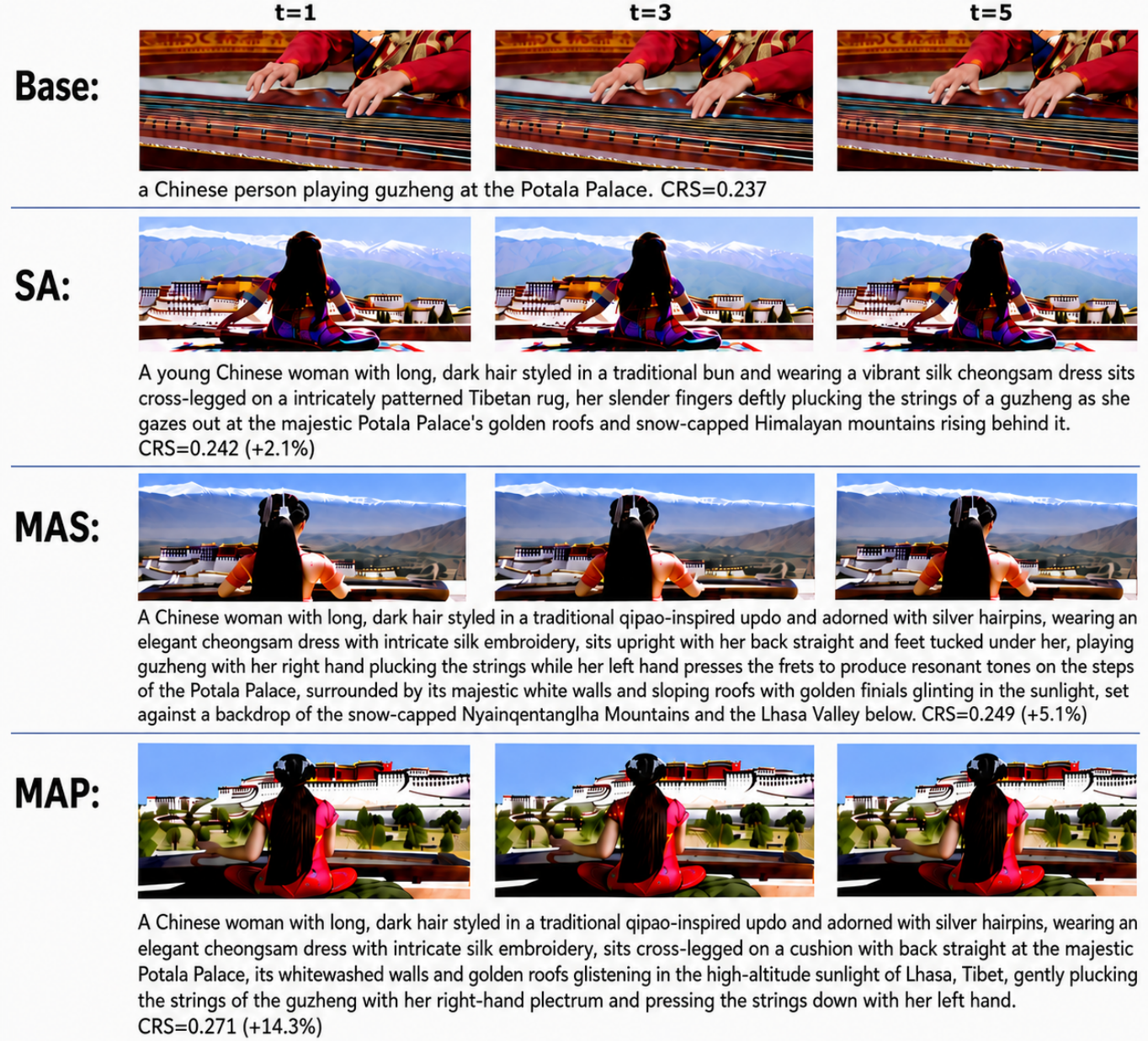}
\caption{Qualitative comparison across all four pipelines (Base, SA, MAS, MAP) for
the prompt \textit{``a Chinese person playing guzheng at the Potala Palace''}
(frames $t=1$, $t=3$, $t=5$). Each pipeline shows a clear progression of cultural
enrichment, with MAP achieving the most balanced and detailed rendering across
person, action, and location dimensions (CRS: Base 0.237 $\to$ SA 0.242 $\to$
MAS 0.249 $\to$ MAP 0.271, +14.3\%).}
\label{fig:qualitative_main_full}
\end{figure*}

\begin{figure*}[!t]
\centering
\includegraphics[width=0.768\textwidth]{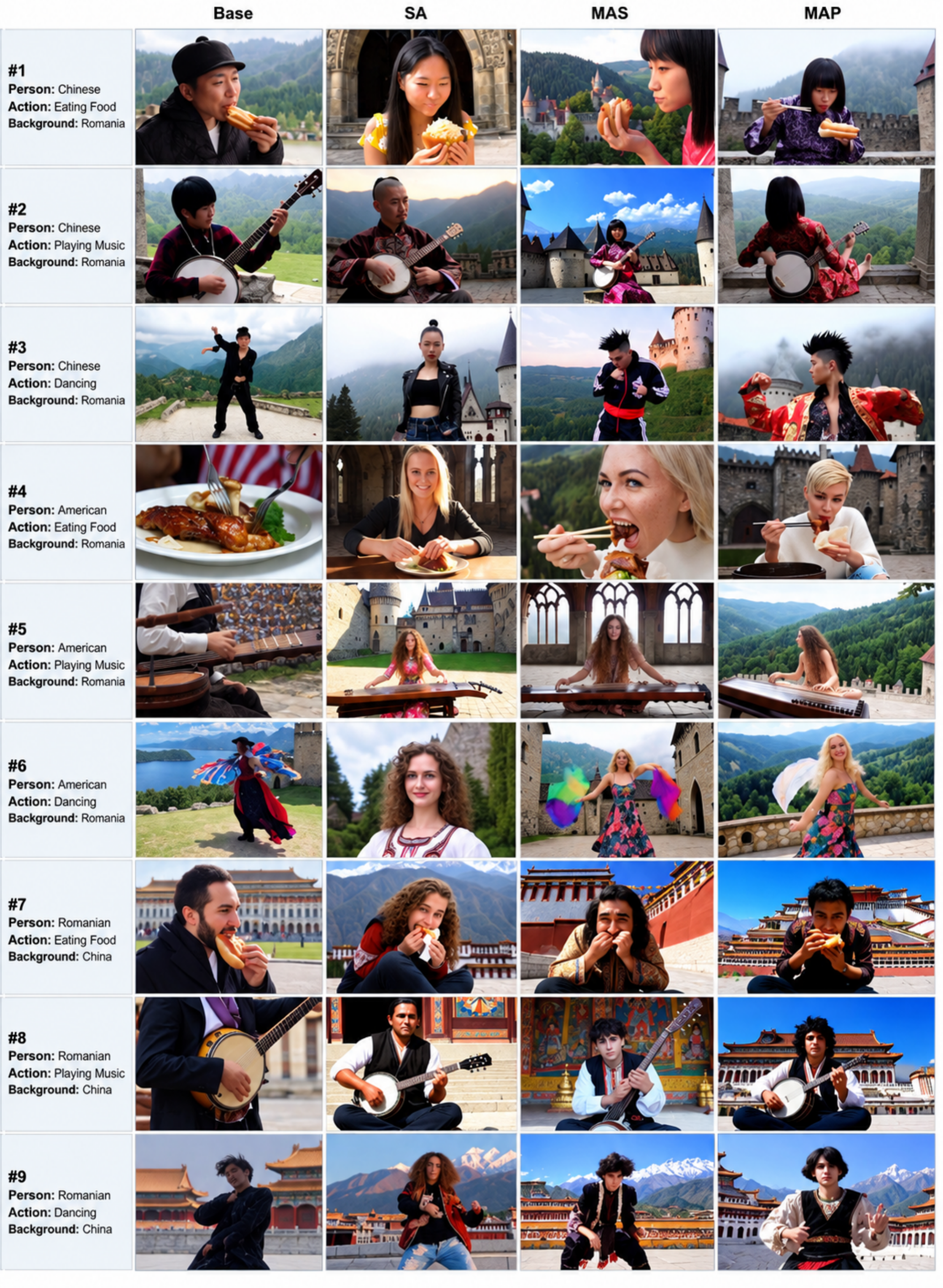}
\caption{Representative cross-cultural qualitative examples across all four
pipelines (Base, SA, MAS, MAP). Each row combines a person, action, and background
drawn from different cultural sources, spanning Chinese, American, and Romanian
settings. The examples illustrate how the refinement pipelines affect cultural
composition across food, music, and dance scenarios.}
\label{fig:cross_cultural_qualitative}
\end{figure*}


\subsection{Qualitative Analysis}
Qualitative results are consistent with quantitative trends. Figure~\ref{fig:qualitative_main_full}
shows a comparison across all four pipelines for the mono-cultural example ``a Chinese
person playing guzheng at the Potala Palace''. Figure~\ref{fig:cross_cultural_qualitative}
further presents representative cross-cultural examples spanning different combinations
of person, action, and background across Chinese, American, and Romanian cultures.
The full five-frame version of the example in Figure~\ref{fig:qualitative_main_full}
is provided in Appendix~\ref{sec:appendix_qualitative} Figure~\ref{fig:qualitative_guzheng},
and further examples can be explored in our dataset.

\paragraph{Pipeline progression and sources of improvement.}
The four pipelines show a clear progression of cultural detail. \textbf{Base} uses 
the original prompt with no enrichment (CRS = 0.237). \textbf{SA} adds basic cues 
across all dimensions -- traditional hair styling, a Tibetan rug, and the Palace's 
golden roofs (CRS = 0.242, +2.1\%). \textbf{MAS} deepens all dimensions further: a 
qipao-inspired updo, specific body posture and hand technique, and a comprehensive 
architectural description (CRS = 0.249, +5.1\%). \textbf{MAP} achieves the most 
comprehensive enrichment across all three dimensions simultaneously (CRS = 0.271, 
+14.3\%), owing to \emph{dimension-specific depth} (each specialized agent brings 
focused expertise a single agent cannot match) and \emph{cross-dimensional balance} (parallel processing gives all three dimensions equally deep enrichment).

\section{Lessons Learned}

\paragraph{Parallel specialization outperforms sequential and single-agent refinement.}
Across a controlled benchmark spanning three cultures and multiple action types,
parallel multi-agent refinement (MAP) achieves the highest cultural relevance,
improving CRS by 4.6\% over Base and 1.6\% over SA, and VLM\_CRS by 38.3\% over Base
and 5.6\% over SA. The largest gain is on location, where LCRS improves by 12.7\%
over Base and VLM\_LCRS rises from 2.86 to 4.43 (+54.9\%). MAP also delivers the most
balanced gains in visual quality (VQ) and temporal consistency (TC), reaching
VQ 61.37 and TC 58.18 versus 60.69 and 52.82 for Base: SA leads on VQ (61.99) and
MAS leads on TC (58.81), but neither matches MAP on both.

\paragraph{Cross-cultural generation benefits most from explicit decomposition.}
Cross-cultural prompts remain substantially more challenging than mono-cultural ones,
yet benefit disproportionately from agent-based refinement. This indicates that explicit
decomposition and recomposition of cultural dimensions is a promising strategy for
cross-cultural generation, and suggests that future systems combining multiple cultural
sources within a single scene may benefit from similarly compositional approaches.

\paragraph{Automatic metrics underestimate culturally nuanced improvements.}
While CLIP-based metrics correlate reasonably with VLM judgments, they systematically
underestimate fine-grained cultural nuances introduced by agent refinement, pointing
to the need for richer semantic reasoning in both T2V systems and their evaluation.

\paragraph{Structured specialization is beneficial in our controlled setting.}
Distributing refinement across culturally specialized agents (MAP) outperforms a single general-purpose agent (SA) on cultural relevance under the same underlying model. These results suggest that dimension-specific decomposition can benefit cultural prompt refinement in our experimental setting, while evaluation across additional LLMs and T2V models is needed before drawing broader conclusions about architecture or model scale.

\section{Conclusion}

In this paper, we introduce multicultural text-to-video generation as a new task,
evaluating how models depict people, actions, and locations from different cultural
backgrounds within a single video. We release the first benchmark for this setting:
243 prompts and 972 videos spanning three cultures, three action categories, and both
mono- and cross-cultural scenarios. Our evaluations show that cross-cultural
composition remains substantially harder than mono-cultural generation, and that
automatic metrics can underestimate the cultural nuance introduced by prompt
refinement. As one strategy for improving cultural fidelity, \textsc{MAVEN} decomposes
prompts into person, action, and location dimensions, each assigned to a culturally
specialized agent, narrowing this gap and yielding more balanced gains than
single-agent or sequential refinement. As generative video becomes an increasingly influential medium for cultural storytelling, education, and marketing, we see this work as a step toward systems that measure and reduce the risk of cultural misrepresentation, helping generative video represent the world's diversity more faithfully. Dataset and code: \url{https://github.com/AIM-SCU/MAVEN}. We encourage future work extending our findings to additional settings, cultures, and languages.

\section*{Limitations}

\paragraph{Limited cultural and activity coverage.}
Our study focuses on three cultures (Chinese, American, Romanian) and three activity categories (food, music, dance), which represent only a small subset of global cultural diversity. These choices were driven by practical considerations, including availability of culturally grounded visual resources and clarity of visual representation. Many cultural dimensions such as gesture, social interaction, spatial organization, color symbolism, and abstract concepts like values or social norms are not explicitly modeled or evaluated. As a result, our findings should not be interpreted as comprehensive across cultures or forms of cultural expression. Future work should expand coverage to a broader range of cultures and activities and incorporate additional cultural dimensions to test the scalability and generality of multi-agent refinement. Such expansion may also surface culture-specific challenges that are not visible in the current setting.

\paragraph{Evaluation on a single text-to-video model.}
All experiments are conducted using a single open-source T2V model CogVideoX-5B \cite{DBLP:conf/iclr/YangTZ00XYHZFYZ25} to ensure controlled comparison. While this isolates the effect of agent-based prompt refinement, different T2V models may exhibit varying sensitivity to culturally enriched prompts due to differences in training data and generation mechanisms. Evaluating the proposed framework across multiple T2V architectures is an important direction for future work to assess model-agnosticity and identify potential model-specific adaptations.

\paragraph{Model-level generation artifacts.}
We observe that CogVideoX-5B \cite{DBLP:conf/iclr/YangTZ00XYHZFYZ25} frequently generates subjects from behind or at oblique angles, obscuring culturally specific facial features and clothing details. This tendency limits the model's ability to render person-level cultural cues and likely contributes to the persistently lower PCRS scores observed across all pipelines. Future work should investigate generation strategies that encourage frontal or culturally expressive subject framing.

\paragraph{Human evaluation.}
Our current evaluation combines CLIP-based metrics with Gemini 2.5 Pro as a VLM judge, providing scalable assessments of cultural relevance across the full benchmark. Human judgments from culturally knowledgeable annotators would provide an important complementary perspective, particularly for nuanced or context-dependent cultural cues that automatic metrics may not fully capture. We therefore interpret our results within the scope of the automatic evaluation framework used in this study and encourage future work to validate these findings with culturally diverse human annotators.

\section*{Ethical Considerations}
Our goal is to improve cultural fidelity and reduce misrepresentation in generative video systems. However, culturally grounded generation also carries risks, including stereotyping or overgeneralization if cultural signals are treated as fixed or homogeneous. The underlying LLMs and VLMs also inherit strong WEIRD \cite{Mihalcea_Ignat_Bai_Borah_Chiruzzo_Jin_Kwizera_Nwatu_Poria_Solorio_2025} biases from their training data \cite{Atari2023WhichHumans} and can amplify representational harms when coverage of under-represented cultures is sparse \cite{DBLP:conf/fat/BenderGMS21}, risks that culturally specialized agents may inherit or magnify. To mitigate these effects, we represent each culture through multiple distinct items rather than a single canonical referent, keep all prompts, refined outputs, and evaluation prompts inspectable, and treat our cultural representations as illustrative rather than exhaustive. We encourage future research to involve broader cultural perspectives and human evaluation to support more responsible and inclusive deployment.

\bibliography{custom}
\bibliographystyle{acl_natbib}

\newpage
\appendix
\section{Reproducibility Details: Compute and Runtime}
\label{sec:appendix_compute}

Experiments are conducted on NVIDIA H100 GPUs.
Generating a single 5-second video takes approximately 3 minutes.
SA and MAP pipelines require one additional minute for prompt refinement, while MAS
requires approximately 2 additional minutes due to its sequential design.
Across all prompts and pipelines, total runtime is approximately 65 GPU hours.
The MAP runtime is comparable to SA because the specialist agents are executed concurrently.
\section{Qualitative Analysis}
\label{sec:appendix_qualitative}

\begin{figure*}[h]
\centering
\includegraphics[width=\textwidth]{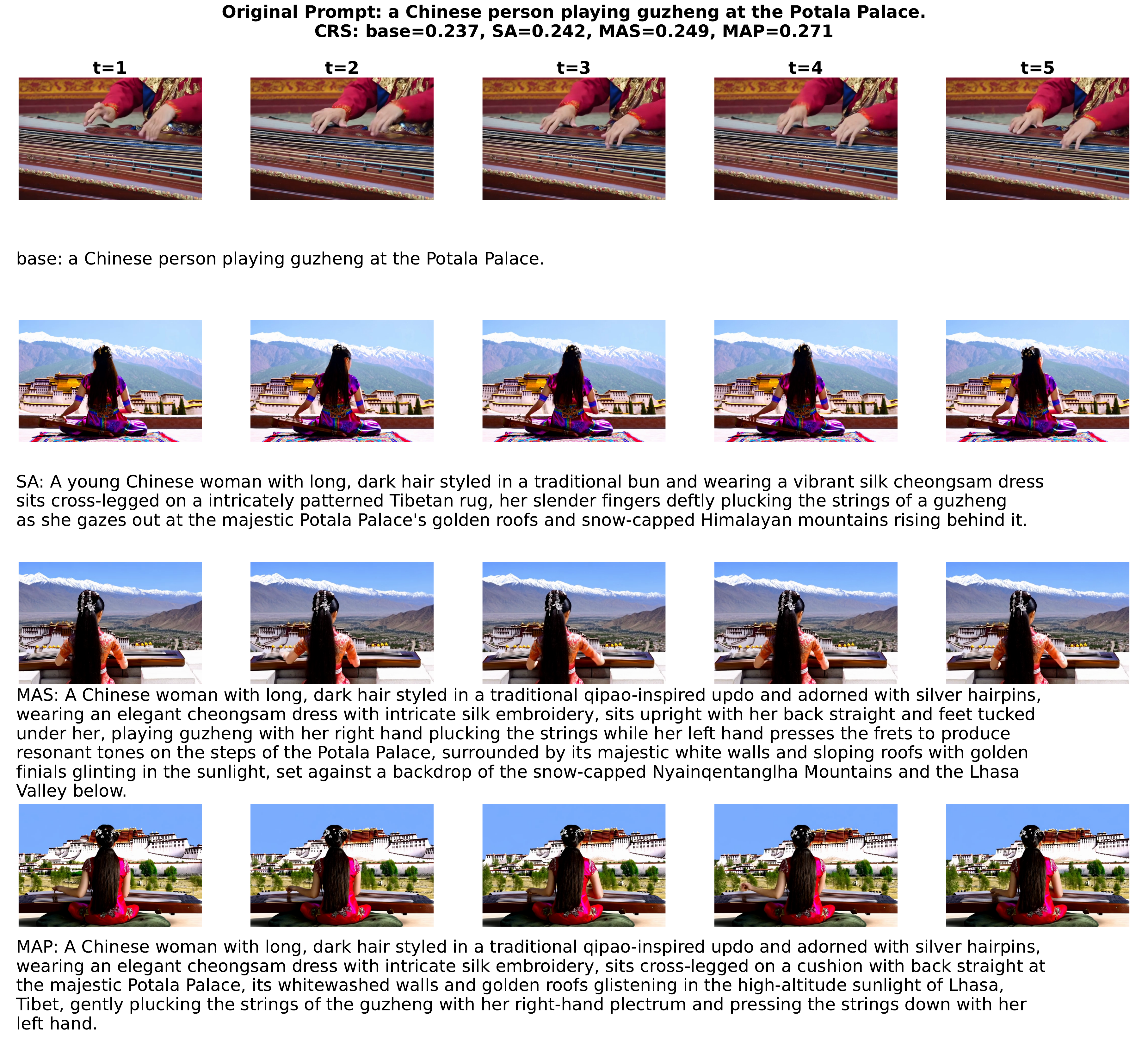}
\caption{Qualitative comparison for a mono-cultural example (``a Chinese person playing guzheng at the Potala Palace''). Five temporal frames are shown for each pipeline (base, SA, MAS, MAP). MAP achieves the highest cultural relevance (CRS: base 0.237, SA 0.242, MAS 0.249, MAP 0.271).}
\label{fig:qualitative_guzheng}
\end{figure*}

\section{Agent System Prompts and Personas}
\label{sec:appendix_prompts}

\subsection{Persona Prompts}
\label{subsec:appendix_personas}

Table~\ref{tab:persona_prompts} presents the persona prompts dynamically generated for each agent based on the cultural dimension and culture source.

\begin{table*}[htbp]
\centering
\small
\begin{tabular}{|l|p{8cm}|}
\hline
\textbf{Agent Type} & \textbf{Persona Template} \\
\hline\hline
PersonAgent & You are a \{culture\} individual who understands typical appearance traits from this culture. \\
\hline
ActionAgent & You are a \{culture\} observer skilled at describing how people typically eat, play music, and dance. \\
\hline
LocationAgent & You are a \{culture\} tour guide who knows how to visually describe iconic landmarks. \\
\hline
SingleShotAgent (mono-cultural) & You are someone familiar with \{culture\} cultural settings and can enrich any part of the scene. \\
\hline
SingleShotAgent (cross-cultural) & You are someone familiar with \{culture1\}, \{culture2\} and \{culture3\} cultural settings and can enrich any part of the scene. \\
\hline
FuseAgent & No culture-specific persona (fusion task only) \\
\hline
\end{tabular}
\caption{Persona prompts for different agent types. The \{culture\} placeholder is dynamically replaced with the specific culture (Chinese, American, or Romanian) based on the dimension being refined. For cross-cultural prompts in SingleShotAgent, all three cultures are listed.}
\label{tab:persona_prompts}
\end{table*}

\subsection{Instruction Prompts for Single-Agent Pipeline}
\label{subsec:appendix_sa_instructions}

Table~\ref{tab:sa_instruction} shows the complete instruction prompt used by the SingleShotAgent for refining all dimensions (person, action, location) simultaneously.

\begin{table*}[htbp]
\centering
\small
\begin{tabular}{|p{12cm}|}
\hline
\textbf{SingleShotAgent Instruction Prompt} \\
\hline\hline
Your task is to improve the full prompt by refining the person, action, and location. \\
\\
Focus on: \\
\hspace{1em}• appearance details (e.g., facial features, hairstyle, clothing) \\
\hspace{1em}• how the action is performed (e.g., body posture, hand movement, visible objects) \\
\hspace{1em}• iconic aspects of the place (e.g., landscape, architecture, lighting, environmental elements) \\
\\
Do not introduce new people, actions, or locations. \\
\\
Return JSON with: \\
\hspace{1em}"refined\_prompt": the updated sentence \\
\hspace{1em}"justification": a brief explanation of how your changes improve the visual clarity or cultural specificity of the scene. \\
\hline
\end{tabular}
\caption{Instruction prompt for SingleShotAgent (SA) pipeline. The agent refines all three dimensions in a single pass.}
\label{tab:sa_instruction}
\end{table*}

\subsection{Instruction Prompts for Multi-Agent Pipelines}
\label{subsec:appendix_ma_instructions}

Tables~\ref{tab:person_instruction}, \ref{tab:action_instruction}, \ref{tab:location_instruction}, and \ref{tab:fuse_instruction} present the instruction prompts for the specialized agents used in multi-agent pipelines (MAS and MAP).

\begin{table*}[htbp]
\centering
\small
\begin{tabular}{|p{12cm}|}
\hline
\textbf{PersonAgent Instruction Prompt} \\
\hline\hline
Your task is to improve only the appearance of the person in the prompt. \\
\\
Focus on visual traits like facial features, hairstyle, or clothing. \\
\\
Do not change the action or the location. \\
\\
Return JSON with: \\
\hspace{1em}"refined\_prompt": the updated sentence \\
\hspace{1em}"justification": a brief explanation of how your addition makes the person more visually distinctive or culturally aligned. \\
\hline
\end{tabular}
\caption{Instruction prompt for PersonAgent. This agent refines only the person dimension.}
\label{tab:person_instruction}
\end{table*}
 
\begin{table*}[htbp]
\centering
\small
\begin{tabular}{|p{12cm}|}
\hline
\textbf{ActionAgent Instruction Prompt} \\
\hline\hline
Your task is to improve only the action portion of the prompt. \\
\\
Focus on how the action is performed: body posture, hand movement, or any visible objects. \\
\\
Do not change the person or the location. \\
\\
Return JSON with: \\
\hspace{1em}"refined\_prompt": the updated sentence \\
\hspace{1em}"justification": a brief explanation of why your added details enhance the clarity, vividness, or cultural alignment of the action. \\
\hline
\end{tabular}
\caption{Instruction prompt for ActionAgent. This agent refines only the action dimension.}
\label{tab:action_instruction}
\end{table*}

\begin{table*}[htbp]
\centering
\small
\begin{tabular}{|p{12cm}|}
\hline
\textbf{LocationAgent Instruction Prompt} \\
\hline\hline
Your task is to improve only the location part of the prompt. \\
\\
Add iconic visual details that make the place recognizable (e.g., landscape, architecture, lighting, environmental elements). \\
\\
Do not change the person or the action. \\
\\
Return JSON with: \\
\hspace{1em}"refined\_prompt": the updated sentence \\
\hspace{1em}"justification": a brief explanation of how your addition makes the location more vivid or culturally recognizable. \\
\hline
\end{tabular}
\caption{Instruction prompt for LocationAgent. This agent refines only the location dimension.}
\label{tab:location_instruction}
\end{table*}

\begin{table*}[htbp]
\centering
\small
\begin{tabular}{|p{12cm}|}
\hline
\textbf{FuseAgent Instruction Prompt} \\
\hline\hline
Your task is to merge three versions of the same scene into one sentence. \\
\\
Keep all appearance, action, and location details. Make the result vivid, coherent, and natural-sounding. \\
\\
Do not add new people, actions, or places. \\
\\
Return JSON with: \\
\hspace{1em}"refined\_prompt": the fused sentence \\
\hspace{1em}"justification": a brief explanation of how you combined the three inputs (e.g., merged phrasing, reordered elements, kept best details). \\
\hline
\end{tabular}
\caption{Instruction prompt for FuseAgent. This agent fuses the three independently refined prompts from PersonAgent, ActionAgent, and LocationAgent into one coherent prompt.}
\label{tab:fuse_instruction}
\end{table*}

\section{VLM Evaluation Prompts}
\label{subsec:appendix_vlm_prompts}

This appendix gives the verbatim prompts used for VLM-as-judge evaluation (Section~\ref{subsec:vlm_eval}). Tables~\ref{tab:vlm_cultural_relevance}, \ref{tab:vlm_visual_similarity}, and~\ref{tab:vlm_text_image_alignment} present the prompts for cultural relevance, visual similarity, and text--image alignment, respectively.

\begin{table*}[!t]
\centering
\small
\begin{tabular}{|p{12cm}|}
\hline
\textbf{VLM Cultural Relevance Evaluation Prompt} \\
\hline\hline
You will be given one image and four culturally annotated sentences describing different aspects of it (overall scene, person, action, and location). \\
\\
Your task is to evaluate how culturally aligned the image is with respect to each sentence — that is, whether the visual features in the image reflect the cultural cues or identities expressed in the text. \\
\\
Specifically, consider appearance, clothing, architecture, symbols, traditions, or any other visual cues linked to a specific country or culture. \\
\\
For each sentence, reason step by step and assign a score between 1 and 5, where: \\
\\
\hspace{1em}1 = The image is not culturally relevant to the description \\
\hspace{1em}5 = The image is highly culturally relevant to the description \\
\\
image: [image\_path] \\
text\_overall: "[overall cultural grounding statement]" \\
text\_person: "[person cultural grounding statement]" \\
text\_action: "[action cultural grounding statement]" \\
text\_location: "[location cultural grounding statement]" \\
\\
The output should be a JSON object ONLY with the following format: \\
\{ \\
\hspace{1em}"overall\_reasoning": "...", \\
\hspace{1em}"overall\_score": number, \\
\hspace{1em}"person\_reasoning": "...", \\
\hspace{1em}"person\_score": number, \\
\hspace{1em}"action\_reasoning": "...", \\
\hspace{1em}"action\_score": number, \\
\hspace{1em}"location\_reasoning": "...", \\
\hspace{1em}"location\_score": number \\
\} \\
\hline
\end{tabular}
\caption{VLM evaluation prompt for cultural relevance. The VLM evaluates how culturally aligned the video frame is with respect to four cultural grounding statements.}
\label{tab:vlm_cultural_relevance}
\end{table*}

\begin{table*}[htbp]
\centering
\small
\begin{tabular}{|p{12cm}|}
\hline
\textbf{VLM Visual Similarity Evaluation Prompt} \\
\hline\hline
You will be shown two images. Your task is to evaluate how visually similar the second image is compared to the first image. \\
\\
Consider all visible aspects, including the person (appearance, pose, clothing), the action (what is happening), the location (environment, background), and any cultural or stylistic details. \\
\\
Please explain your reasoning step by step, and then assign a score between 1 and 5: \\
\\
\hspace{1em}1 = Very different visually \\
\hspace{1em}5 = Nearly identical visually \\
\\
image\_1: (reference image) \\
image\_2: (comparison image) \\
\\
The output should be a JSON object ONLY with the following format: \\
\{ \\
\hspace{1em}"reasoning": "...", \\
\hspace{1em}"score": number \\
\} \\
\hline
\end{tabular}
\caption{VLM evaluation prompt for visual similarity. The VLM compares the visual similarity between base video frames and agent-refined video frames.}
\label{tab:vlm_visual_similarity}
\end{table*}

\begin{table*}[htbp]
\centering
\small
\begin{tabular}{|p{12cm}|}
\hline
\textbf{VLM Text-Image Alignment Evaluation Prompt} \\
\hline\hline
You will be shown one image and multiple sentences that describe what the image is expected to show. \\
\\
Your task is to evaluate how well the visual content of the image aligns with each description. \\
\\
Focus on whether the image clearly reflects the elements described — such as people, actions, locations, objects, or any culturally or visually specific features. \\
\\
For each sentence, please explain your reasoning step by step, and assign a score between 1 and 5, where: \\
\\
\hspace{1em}1 = The image does not match the description \\
\hspace{1em}5 = The image clearly and fully matches the description \\
\\
Please evaluate all sentences and return your response as a JSON object with this exact format: \\
\{ \\
\hspace{1em}"text1\_reasoning": "...", \\
\hspace{1em}"text1\_score": number, \\
\hspace{1em}"text2\_reasoning": "...", \\
\hspace{1em}"text2\_score": number, \\
\hspace{1em}"text3\_reasoning": "...", \\
\hspace{1em}"text3\_score": number, \\
\hspace{1em}"text4\_reasoning": "...", \\
\hspace{1em}"text4\_score": number, \\
\hspace{1em}"text5\_reasoning": "...", \\
\hspace{1em}"text5\_score": number, \\
\hspace{1em}"text6\_reasoning": "...", \\
\hspace{1em}"text6\_score": number \\
\} \\
\\
Note: text1-text4 are the four cultural grounding statements (overall, person, action, location), text5 is the original prompt, and text6 is the final refined prompt. \\
\hline
\end{tabular}
\caption{VLM evaluation prompt for text-image alignment. The VLM evaluates the alignment between video frames and multiple text descriptions including cultural grounding statements and prompts.}
\label{tab:vlm_text_image_alignment}
\end{table*}



\end{document}